\pdfoutput=1
\documentclass{article}

\PassOptionsToPackage{numbers}{natbib}
 
\usepackage[preprint]{nips_2018}

\usepackage[utf8]{inputenc} 
\usepackage[T1]{fontenc}    
\usepackage{booktabs}       
\usepackage{amsfonts}       
\usepackage{nicefrac}       
\usepackage{microtype}      
\usepackage[caption = true]{subfig}

\usepackage[]{mathtools}
\usepackage{amsthm}

\newcommand{\figref}[1]{Figure \ref{#1}}
\newcommand{\secref}[1]{Section \ref{#1}}

\newtheorem{theorem}{Theorem}

\title{Unbiased Estimation of the Value of an Optimized Policy}

\author{
  Elon Portugaly\\
  Bing Ads \\
  \texttt{elonp@microsoft.com}
  \And
  Joseph J. Pfeiffer III\\
  Bing Ads \\
  \texttt{joelpf@microsoft.com}
}

\date{\today}
\begin{document}
\setlength{\abovedisplayskip}{2pt}
\setlength{\belowdisplayskip}{2pt}
\setlength{\belowcaptionskip}{-20pt}

\maketitle

\begin{abstract}
    Randomized trials, also known as A/B tests, are used to select between two policies: a control and a treatment.  Given a corresponding set of features, we can ideally learn an optimized policy P that maps the A/B test data features to action space and optimizes reward.  However, although A/B testing provides an unbiased estimator for the value of deploying B (i.e., switching from policy A to B), direct application of those samples to learn the the optimized policy P generally does not provide an unbiased estimator of the value of P as the samples were observed when constructing P. In situations where the cost and risks associated of deploying a policy are high, such an unbiased estimator is highly desirable.
    
    We present a procedure for learning optimized policies and getting unbiased estimates for the value of deploying them.  We wrap any policy learning procedure with a bagging process and obtain out-of-bag policy inclusion decisions for each sample.  We then prove that inverse-propensity-weighting effect estimator is unbiased when applied to the optimized subset.  Likewise, we apply the same idea to obtain out-of-bag unbiased per-sample value estimate of the measurement that is independent of the randomized treatment, and use these estimates to build an unbiased doubly-robust effect estimator.  Lastly, we empirically shown that even when the average treatment effect is negative we can find a positive optimized policy.

\end{abstract}

\section{Introduction}

Randomized trials, also known as A/B tests, are used to select between two policies: \emph{A} which assigns the control action to all samples, and \emph{B} which assigns the treatment action to all samples.  For example, in medical trials a drug treatment (\emph{B}) might perform, on average, worse than no action (\emph{A}), resulting in the A/B test choosing to reject the drug.  As proposed by \cite{li2010contextual,swaminathan2015batch,wager2017estimation}, a more refined approach could use the same data to learn an optimized policy \emph{$\Pi$} mapping sample features to action, optimizing reward.  For the medical trial example, the optimized policy \emph{$\Pi$} might choose a subset of the population to give the drug treatment based on features (e.g., age), while not applying the treatment to the remaining population.

Note, however, that while A/B testing provides an unbiased estimate for the value of deploying \emph{B} (i.e. switching from policy \emph{A} to policy \emph{B}), such an estimate is generally not available for deploying an optimized policy \emph{$\Pi$}, because the samples were already observed when constructing \emph{$\Pi$}.  In situations where the cost and risks associated of deploying a policy are high, such an unbiased estimator is highly desirable.

We present a procedure for learning optimized policies and getting unbiased estimates for the value of deploying them.  We wrap any policy learning procedure with a bagging process \cite{breiman1996bagging}, we then proceed to obtain out-of-bag policy inclusion decisions for each of the samples.  For each sample, the out-of-bag effect policy decision is not a function of that sample's treatment indicator or measurement, and is therefore independent of the randomized treatment.  We can therefore show that the inverse-propensity-weighting effect estimator \cite{rosenbaum1983central,lunceford2004stratification} is unbiased when applied to the optimized subset.  Furthermore, we apply the same idea - bagging followed by out-of-bag prediction - to any regression learning procedure that ignores the treatment, and get a per-sample estimate of the measurement that is indepedent of the randomized treatment.  This can be plugged in to a doubly-robust effect estimator \cite{dudik2011doubly,cassel1976some,robins1994estimation,robins1995semiparametric,lunceford2004stratification,kang2007demystifying}, which will also produce an unbiased average effect estimate when applied to the optimized subset.

The rest of the paper proceeds as follows: Section \ref{sec:theory} introduces the notation and theoretical framework used, then describes and proves sufficient conditions for providing an unbiased estimation for the value of an optimized policy, and finally proves that the bagging procedure described above meets these conditions.  Section \ref{sec:simulations} describes simulation studies evaluating the methods described herein.  We simulate experimental data by taking standard regression datasets and adding synthetic heterogeneous treatment effects to them.  We then apply several policy learning procedures to the simulated data, and estimate the value of the learned policy.  We show that the estimated value is indeed an unbiased estimate of the synthetic treatment effect on the learned policy, and study the power of the different learners to find optimal policies.  Section \ref{sec:real} describes the application the same methods to real world trials, and illustrates the power to detect positive policies even when the average treatment effect is negative.  We conclude in \secref{sec:conclusion}.

\section{Related Work}
\label{sec:relatedwork}

The challenges and benefits of A/B testing are explained in detail in \cite{kohavi2013online}. The doubly-robust estimator has been described in \cite{cassel1976some, robins1994estimation, robins1995semiparametric, lunceford2004stratification}, and more recently, in the context of policy evaluation by \cite{dudik2011doubly}.  It is used by \cite{poyarkov2016boosted} to reduce the variance of A/B testing experiments.  However, their boosted decision trees are trained on the same data they generate predictions for, without taking any measures to ensure the prediction for sample $i$ is not a function of the measurement for that same sample.  As warned by \cite{deng2013improving}, this will lead to an underestimate of the treatment effect, and in the case of tree-based predictions, the bias is likely to be strong, since every sample is guaranteed to be mapped to the same leaf it was mapped to in training time.

\cite{li2010contextual} optimize policies for news recommendation applying a contextual-bandit algorithm to randomized data, whereas \cite{swaminathan2015batch} Provide a framework for policy optimization given data from a randomized experiment (or logged bandit data), and provide proven bounds on the value of their optimized policy.

\cite{athey2016recursive} and \cite{wager2017estimation} describe different ways to construct decision trees and random forest for heterogeneous effect estimation, i.e. for estimating the effect on a sample given its features.  Most of our effect estimators are implementations of algorithms described there.  They also introduce the idea of \textit{honest} prediction, where the predicted associated with a tree leaf is a function of samples that were not used in training that tree.  This is related, but not identical, to our use of out-of-bag predictions.  They use the \textit{honest} characteristic to prove that their models provide consistent (asymptotically unbiased) estimations, whereas we use the out-of-bag predictions to ensure independence of the prediction and the sample's treatment, thus enabling a second estimator to be unbiased.

\section{Theoretical analysis}
\label{sec:theory}

Following the Rubin Causal Model \cite{rubin1974estimating,imbens_rubin_2015_c1}:

Let $N$ be the number of experimentation units and $i, j \in \left[N\right]$ be indices into those units.  We let $X_i \in \mathcal{X}$ be fixed and observed features of the experimentation units and $y_i^{\left(1\right)},y_i^{\left(0\right)}\in \mathcal{R}$ be fixed and unobserved potential outcomes for treatment (1) and control (0), respectively.
We let $\tau_i \equiv y_i^{\left(1\right)} - y_i^{\left(0\right)}$ denote the unit treatment effect.  The treatment indicators, $T_i$, are independent Bernoulli random variables, with $P\left(T_i=1\right)=p_i$, for some fixed $p_i$.  The observed outcomes are denoted by $Y_i = y_i^{\left(T_i\right)}$.

We will focus on determinstic policies.  A policy $\Pi:\mathcal{X} \rightarrow \{0,1\}$ is a mapping from feature space to treatment decision.  In the context of evaluating the policy on our fixed set of experimentation units, the policy can be considered as a subset of the indices: $\Pi = \left\{i \mid \Pi\left(X_i\right)=1 \right\}$.  We will also use the notation $\Pi_i=\Pi\left(X_i\right)$.

The (unknown) value of (deploying) policy $\Pi$, i.e. switching from assigning all units to control (0) to assigning according to $\Pi$ is:
\begin{equation}
\label{eqn:value-of-a-policy}
V_\Pi = \sum_i y_i^{\left(\Pi_i\right)} - \sum_i y_i^{\left(0\right)} = \sum_{i \in \Pi}  \tau_i
\end{equation}

The value of $\Pi$ is unknown because it depends on $y_i^{\left(\Pi_i\right)}$, which is only observed in for a subset of the units.

\subsection{Sufficient conditions for the unbiased estimation of the value of an optimized policy}

Let $\Pi$ be an optimized policy, i.e. $\Pi$ is a function of $\left\{X_i,T_i,Y_i\right\}_i$, and via them a function of the random variables $\left\{T_i\right\}_i$ and therefore a random variable in itself.

Let $\tilde{y_i}^{\left(0\right)}$ and $\tilde{y_i}^{\left(1\right)}$ be predictors for $y_i^{\left(0\right)}$ and $y_i^{\left(1\right)}$, respectively.

Let $\hat{V}\left(\Pi\right)$ be the doubly-robust estimator for the value of $\Pi$:

\begin{equation}
\label{eqn:doubly-robust}
\hat{V}\left(\Pi\right) = \sum_{i \in \Pi} \left[\frac{T_i}{p_i} \left(Y_i -\tilde{y_i}^{\left(1\right)}\right) - \frac{1-T_i}{1-p_i} \left(Y_i -\tilde{y_i}^{\left(0\right)}\right) + \tilde{y_i}^{\left(1\right)} - \tilde{y_i}^{\left(0\right)}\right]
\end{equation}

The inverse-propensity-weighting estimator is unbiased since it is a special case of the doubly-robust estimator with $\tilde{y_i}^{\left(\cdot\right)}=0$.

As shown by \cite{rosenbaum1983central,lunceford2004stratification,dudik2011doubly} and others the doubly-robust estimator is an unbiased estimator of $V_\Pi$ if $\Pi$ is fixed.

The following theorem states and proves sufficient conditions for $\hat{V}\left(\Pi\right)$ and $\tilde{V}_\Pi$ to be unbiased when $\Pi$ is optimized:

\begin{theorem}
\label{thrm:unbiasedness}
	The doubly robust estimator $\hat{V}\left(\Pi\right)$ is an unbiased estimator of the value of $\Pi$ if for each sample $i$, both the policy inclusion decision and the predictors are independent of the treatment assigned to sample $i$, i.e.: $T_i \bot \left\{\Pi_i, \tilde{y_i}^{\left(0\right)}, \tilde{y_i}^{\left(1\right)}\right\}$.
\end{theorem}
Note that we do not require $\Pi_i$ to be independent of $\tilde{y_i}^{\left(A\right)}$ and $\tilde{y_i}^{\left(B\right)}$.

\begin{proof}
Recall that the value of any policy is given by \eqref{eqn:value-of-a-policy}.  
This applies for our optimized policy $\Pi$, but note that because $\Pi$ depends on the data, $V_\Pi$ is a random variable and, additionally, it is unobserved.

Let $B_\Pi = \hat{V}\left(\Pi\right) - V_\Pi$ be the difference between our estimator and the true value of $S$.
Again, note that $B_S$ is an unobserved random variable.  We will show that it is zero in expectation, and by that prove that $\hat{V_S}$ is an unbiased estimator of $V_S$.
	
\begin{align*}
	B_\Pi & = \hat{V}\left(\Pi\right) - V_\Pi \\
	&= \sum_{i \in \Pi} \left[\frac{T_i}{p_i} \left(Y_i -\tilde{y_i}^{\left(1\right)}\right) - \frac{1-T_i}{1-p_i} \left(Y_i -\tilde{y_i}^{\left(0\right)}\right) + \tilde{y_i}^{\left(1\right)} - \tilde{y_i}^{\left(0\right)}\right] - \sum_{i \in \Pi}  \tau_i \\
	&= \sum_i \Pi_i\left[\frac{T_i}{p_i} \left(Y_i -\tilde{y_i}^{\left(1\right)}\right) - \frac{1-T_i}{1-p_i} \left(Y_i -\tilde{y_i}^{\left(0\right)}\right) + \tilde{y_i}^{\left(1\right)} - \tilde{y_i}^{\left(0\right)} - \tau_i\right] \\
	\shortintertext{We replace $\tau_i$ and $Y_i$ by their definitions:}
	&= \sum_i \Pi_i\left[\frac{T_i}{p_i} \left(y_i^{\left(1\right)} -\tilde{y_i}^{\left(1\right)}\right) - \frac{1-T_i}{1-p_i} \left(y_i^{\left(0\right)} -\tilde{y_i}^{\left(0\right)}\right) + \tilde{y_i}^{\left(1\right)} - \tilde{y_i}^{\left(0\right)} - y_i^{\left(1\right)} + y_i^{\left(0\right)}\right] \\
	\shortintertext{Replacing $y_i^{\left(T_i\right)}$ with $y_i^{\left(1\right)}$ or $y_i^{\left(0\right)}$ as needed, and rearanging:}
	&= \sum_i \left[ \underbrace{\Pi_i\left(y_i^{\left(1\right)} -\tilde{y_i}^{\left(1\right)}\right)}_{D_i^{\left(1\right)}}\underbrace{\left(\frac{T_i}{p_i}-1\right)}_{W_i^{\left(1\right)}} - \underbrace{\Pi_i \left(y_i^{\left(0\right)} -\tilde{y_i}^{\left(0\right)}\right)}_{D_i^{\left(0\right)}}\underbrace{\left(\frac{1-T_i}{1-p_i}-1\right)}_{W_i^{\left(0\right)}} \right]
\end{align*}

The $W_i^{\left(\cdot\right)}$ translate the treatment indicators into importance weights.  In the fair-coin case, they map $1 \rightarrow 1$ and $0 \rightarrow -1$.  The $D_i^{\left(\cdot\right)}$ terms capture the errors of the invariant predictors, when they matter, i.e. when the sample is in the policy.

On expectation the $W_i^{\left(\cdot\right)}$ are 0:
\begin{align*}
	E\left[W_i^{\left(1\right)}\right] &= E\left[\frac{T_i}{p_i}-1\right] = \frac{E\left[T_i\right]}{p_i}-1 = \frac{p_i}{p_i} - 1 = 0 \\
	E\left[W_i^{\left(0\right)}\right] &= E\left[\frac{1-T_i}{1-p_i}-1\right] = \frac{1-E\left[T_i\right]}{1-p_i}-1 = \frac{1-p_i}{1-p_i} - 1 = 0
\end{align*}
.

Note that since $T_i \bot \left\{Pi_i, \tilde{y_i}^{\left(\cdot\right)} \right\}$ and $y_i^{\left(\cdot\right)}$ are constants, any function of $T_i$ is independent of any function of $\tilde{y_i}^{\left(\cdot\right)}$, $\Pi_i$ and $y_i^{\left(\cdot\right)}$.  In particular, $W_i^{\left(\cdot\right)} \bot D_i^{\left(\cdot\right)}$, and hence from multiplicity of expectation of independent variables, we have:

\begin{align*}
	E\left[W_i D_i\right] = \underbrace{E\left[W_i\right]}_{0}E\left[D_i\right] = 0
\end{align*}

And from linearity of expectation, we conclude that the bias is 0 on expectation:

\begin{align*}
	E\left[B_\Pi\right] = E\left[\sum_i D_i^{\left(1\right)} W_i^{\left(1\right)} - D_i^{\left(0\right)} W_i^{\left(0\right)}\right] = \sum_iE\left[D_i^{\left(1\right)} W_i^{\left(1\right)}\right] - E\left[D_i^{\left(0\right)} W_i^{\left(0\right)}\right] = 0
\end{align*}

\end{proof}

\subsection{Producing treatment-invariant predictors and policy indicators}
\label{sec:effectlearners}

To get $\Pi_i$ and $\tilde{y_i}^{\left(\cdot\right)}$ that are guaranteed to be independent of $T_i$, we will wrap any learning procedure generating these decisions and predictions with a bagging procedure followed by out-of-bag estimations.  The following section provides details.

We will assume access to three training procedures:

Let $L^{\left(\tau\right)}$, $L^{\left(0\right)}$, $L^{\left(1\right)}$ be training procedures attempting to produce predictors for the effect ($\tau$), potential outcome under control ($y^{\left(0\right)}$), and potential outcome under treatment ($y^{\left(1\right)}$) respectively. Each of $L^{\left(\cdot\right)}$ takes a set of samples $\left\{x_k, t_k, y_k\right\}_k$, $x_k \in \mathcal{X}$, $t_k \in \left\{0, 1\right\}$, ${y_k} \in \mathcal{R}$ and outputs a predictor $l : \mathcal{X} \rightarrow \mathcal{R}$.  Note that while the learners have access to treatment indicators, the predictor does not.

As an example, $L^{\left(1\right)}$ could ignore the control samples, and learn a regressor on the basis of the treated samples, $L^{\left(0\right)}$ could do the same for the control samples, and $L^{\left(\tau\right)}$ could produce a predictor that returns the difference between the predictors returned by the former two.  Alternatively, we could choose for $L^{\left(1\right)}$ and $L^{\left(0\right)}$ to be the same process, ignoring the treatment indicators.

Wrap each of learners with a bagging process, and the use out-of-bag predictions to create a policy and a treatment-invariant effect prediction:

Training using bagging:

For $m$ in $[M]$
\begin{itemize}
    \setlength\itemsep{0em}
	\item Let $S_m$ be a set of $N$ units drawn with replacement from $\left\{X_i, t_i, y_i\right\}_i$.
	\item Let $l_m^{\left(0\right)}$, $l_m^{\left(1\right)}$, $l_m^{\left(\tau\right)}$ be the functions learned by $L^{\left(0\right)}$, $L^{\left(1\right)}$, $L^{\left(\tau\right)}$, respectively when given $S_m$.
\end{itemize}

Out-of-bag prediction, for each $i \in \left[ N \right]$
\begin{itemize}
    \setlength\itemsep{0em}
	\item Let $M_i = \left\{ m \mid i \notin S_m \right\}$ be the indicators of predictors that did not use $i$ in their training.
	\item Let $\mathcal{L}_i^{\left(1\right)} = \left\{ l_m^{\left(1\right)}\left(x_i\right) \right\}_{m \in M_i}$ be the set of predictions made by the $l_m^{\left(1\right)}$ predictors indexed by $M_i$, when given the features of sample $i$.
	\item Likewise let $\mathcal{L}_i^{\left(0\right)} = \left\{ l_m^{\left(0\right)}\left(x_i\right) \right\}_{m \in M_i}$ and $\mathcal{L}_i^{\left(\tau\right)} = \left\{ l_m^{\left(\tau\right)}\left(x_i\right) \right\}_{m \in M_i}$
	\item Use some aggregation function(s) $f^{\left(\cdot\right)}$ to take the three sets of predictions and produce two predictors $\tilde{y_i}^{\left(0\right)}=f^{\left(0\right)}\left(\mathcal{L}_m^{\left(\cdot\right)}\right)$, $\tilde{y_i}^{\left(1\right)}=f^{\left(1\right)}\left(\mathcal{L}_m^{\left(1\right)}\right)$ and one policy decision $\Pi_i=f^{\left(\tau\right)}\left(\mathcal{L}_m^{\left(\tau\right)}\right)$
\end{itemize}

For the predictors $\tilde{y_i}^{\left(\cdot\right)}$, the most reasonable aggregation function is the average.

One can consider many options for the aggregation function defining the policy.  Examples include taking the average and thresholding on 0, thresholding each individual prediction and then taking a majority vote, or being conservative and selecting the sample into the policy only if the average is at least two standard deviations about 0.

Note that for each $i$, $\tilde{y_i}^{\left(\cdot\right)}$ and $\Pi_i$ are functions of $\{x_j, T_j, Y_j\}_{j \in N\setminus\left\{i\right\}}$ and of $x_i$, but not of $T_i$ or $Y_i$.  Since the $x_j$ are fixed, $T_j$ is independent of $T_i$ for all $j \neq i$, and $Y_j$ is a deterministic function of $T_j$, we can conclude that $T_i \bot \left\{\Pi_i, \tilde{y_i}^{\left(0\right)}, \tilde{y_i}^{\left(1\right)}\right\}$, as required by Theorem \ref{thrm:unbiasedness}.

\section{Simulation Experiments}
\label{sec:simulations}
In this section we evaluate the proposed policy optimization procedure and provide empirical evidence for the unbiased estimator.  Further, we demonstrate scenarios where failing to utilize an unbiased estimator can produce detrimental results, even in the (unobtainable) case of fully available treatment effect information.  We will first demonstrate our unbiased estimator's effectiveness in a synthetic and controlled augmentation of classic public datasets, followed by an evaluation of a treatment effect on a real world experiment on business customers to a large online service.

\subsection{Learners}
\label{sec:optimizers}

As discussed in \secref{sec:effectlearners}, we require training procedures $L^{\left(\tau\right)}$, $L^{\left(0\right)}$, $L^{\left(1\right)}$ producing predictors for effect ($\tau$), potential outcome under control ($y^{\left(0\right)}$), and potential outcome under treatment ($y^{\left(1\right)}$).  For the potential outcome predictors any standard regressor wrapped with bagging is sufficient: we apply Extremely Randomized Trees \cite{geurts2006ert} with 10,000 trees, and use the same prediction $\tilde{y_i}=\tilde{y_i}^{\left(0\right)}=\tilde{y_i}^{\left(1\right)}$ for both potential outcomes.  In contrast, for estimating the (directly) unobservable effect we study a variety of approaches.  For each bagged sample-set $m \in [M]$, we apply

\begin{itemize}
    \setlength\itemsep{0em}
    \item \textbf{LR:} Train linear regression models $\tilde z^1_m$ and $\tilde z^0_m$ using the treatment samples and control samples respectively.  Then $l_m^{\left(\tau\right)} = \tilde z^1_m - \tilde z^0_m$.
    \item \textbf{DT:} Train regression trees $\tilde z^1_m$ and $\tilde z^0_m$ using the treatment samples and control samples respectively.  Then $l_m^{\left(\tau\right)} = \tilde z^1_m - \tilde z^0_m$.
    \item \textbf{ST:} Train $\tilde z_m : \mathcal X \times \mathcal T \rightarrow \mathcal R$ as a single regression tree trained on all given samples. Then $l_m^{\left(\tau\right)}(x) = \tilde z_m(x, 1) - \tilde z_m(x, 0)$.
    \item \textbf{TN:} Train $\tilde z_m : \mathcal X \rightarrow \mathcal R$ as a single regression tree, trained on all given samples, but without looking at the treatment indicators. Let $\tilde z^0_m(x_i)$ be the \textit{control} samples that share a leaf prediction for sample $s_i$, and $\tilde z^1_m(x_i)$ be the \textit{treatment} samples that share a leaf prediction for sample $i$.  Then $l_m^{\left(\tau\right)} = \frac{1}{\left|\tilde z^1_m\right|}\sum_{j \in \tilde z^1_m} y_j - \frac{1}{\left|\tilde z^0_m\right|}\sum_{j \in \tilde z^0_m} y_j$ is the difference-of-means estimator with the leaf.  We create two approaches based on this technique: the first, \textbf{TN-IB}, uses the set $S_m$ to create the estimate $l_m^{\left(\tau\right)}$, while the second, \textbf{TN-OOB}, uses $S_{\not m} = S \setminus \{S_m, i\}$ to create the estimate $l_m^{\left(\tau\right)}$.
    \item \textbf{TE:} Following \cite{wager2017estimation}, we create a regression tree that directly optimizes prediction of $\tau$ by chosing splits that maximize the variance of predicted $\hat{\tau}$ across the leaves.
\end{itemize}

\textbf{ST}, \textbf{ST}, and \textbf{TE} have been described by \cite{athey2016recursive} and \cite{wager2017estimation}.

For the studies, we wrap the above estimators in 1,000 bags.  We apply two aggregation functions to out-of-bag predictions: average-estimates, denoted \textbf{P$^{\boldsymbol E}$} (for some training procedure \textbf{P}), takes the average of the prediction across the bags, and adds the sample to the policy if the average is above 0; votes, denoted \textbf{P$^{\boldsymbol V}$}, converts each out-of-bag prediction into a vote whether $l_m^{\left(\tau\right)}(x_i)>0$, then decides inclusion in policy by majority voting across the bags.

\subsection{Simulation studies}
Taking a regression benchmark dataset, we draw simulated potential effects from some distribution conditioned on the features, randomly split the samples to treatment and control sets, synthetically modify the targets with the potential effects, and run the policy optimization and evaluation process.  Since the effect is simulated, we have a ground truth against which to compare the predictions.

We repeat the process 1,000 times for several simulated effect distributions and several policy optimizers.  We verify that the effect estimation is unbiased for all effect distributions and policy optimizers.  We also evaluate the performance of the different optimizers.

We utilize the Boston housing dataset \cite{dua2017uci,harrison1978boston} for our synthetic analysis.  This allows for reproducibility of our work through a simple python package, which we will release.

\subsubsection{Simulated Effect Distributions}
\label{sec:simulatedeffectdistributions}
For each dataset we treat the original regression target column as the potential outcome under control, $y^{\left(0\right)}$.  We then simulate a synthetic treatment effect by picking a feature $X_{:j}$ and let $S^+ = \left\{i \in [N] | x_{ij} \text{ is in top quintile of } X_{:j} \right\}$, and setting $\tau_i$ for $i \in N$ according to:

\begin{align*}
\left(\mu_i, \sigma_i\right) &= 
\begin{cases}
	\left(\mu_{+}, \sigma_{+}\right),& \text{if } i \in S^+\\
	\left(\mu_{-}, \sigma_{-}\right),& \text{otherwise}\\
\end{cases} \\
\tau_i &= \mathcal{N}(\mu_i, \sigma_i) \\
y^{\left(1\right)} &= y^{\left(0\right)} + \tau_i
\end{align*}

\noindent where $i \in S^+$ get noisy positive treatment effects and $i \notin S^+$ get noisy negative treatment effects (without loss of generatility, we set $\mu_+ > 0 > \mu_{-}$).  Throughout this section we ground $\mu,\sigma$ as magnitudes with respect to the treatment standard deviation of $y^{\left(0\right)}$, which we denote $\sigma_y$.  For example, denoting $\mu_+=1.5$ is interpreted as $1.5\sigma_y$; in this way we examine the treatment effect as a function of the inherent noise already present in the data.

For each simulation, we assign each sample a treatment $t_i$ by a fair-coin toss, apply the synthetic treatment effect accordingly.  The synthetic dataset presented to each policy optimizer is $\{x_i, t_i, y^{\left(t_i\right)}\}$.

\textbf{Benchmark Policies:} The synthetic setup creates a number of natural comparisons for our effect optimizations to compete with.  To start, we test against two na{\"i}ve policies: selecting no-sample (i.e. policy \emph{A}), that has a constant value of 0, and selecting all samples (i.e. policy \emph{B}), \textbf{TakeAll} bellow: in our simulation this policy is by construction strongly negative.  Additionally, we compare against an \textbf{Oracle} policy that assumes complete information and always selects $S^+$.  Lastly, we compare against a \textbf{NoisyOracle} policy that chooses all samples with a positive $\tau$ (like \textbf{Oracle}, this is also an unrealistic optimizer, since it assumes access to $\tau$).

\textbf{Criteria:}  Our analysis focuses on three measures: Effect Mean ($\sum_{i\in\Pi} \mu_i$), Realized Effect ($\sum_{i\in\Pi} \tau_i$) and Estimated Effect ($\hat{V}\left(\Pi\right)$).  These measures are generally stochastic across permutations, however we are mostly interested in the stochastisity of the estimator bias.  We calculate the standard deviations of the difference between Realized Effect and Effect Mean, and of the difference between Estimated Effect and Effect Mean, provide error bars of 2 standard deviations for those measurements.

\begin{figure}
    \centering
    \subfloat[]{\includegraphics[width=0.3\textwidth]{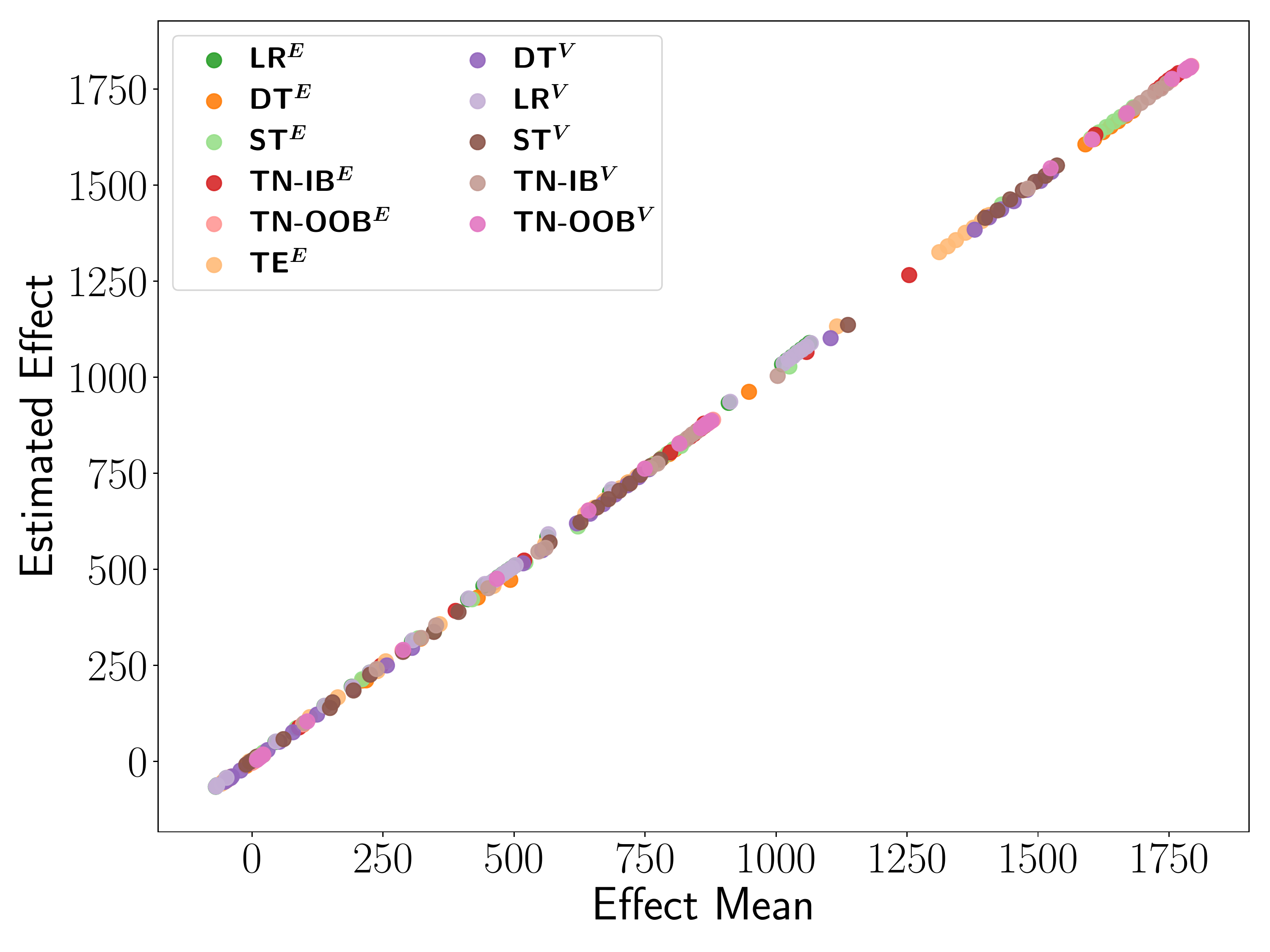}}
    \subfloat[]{\includegraphics[width=0.3\textwidth]{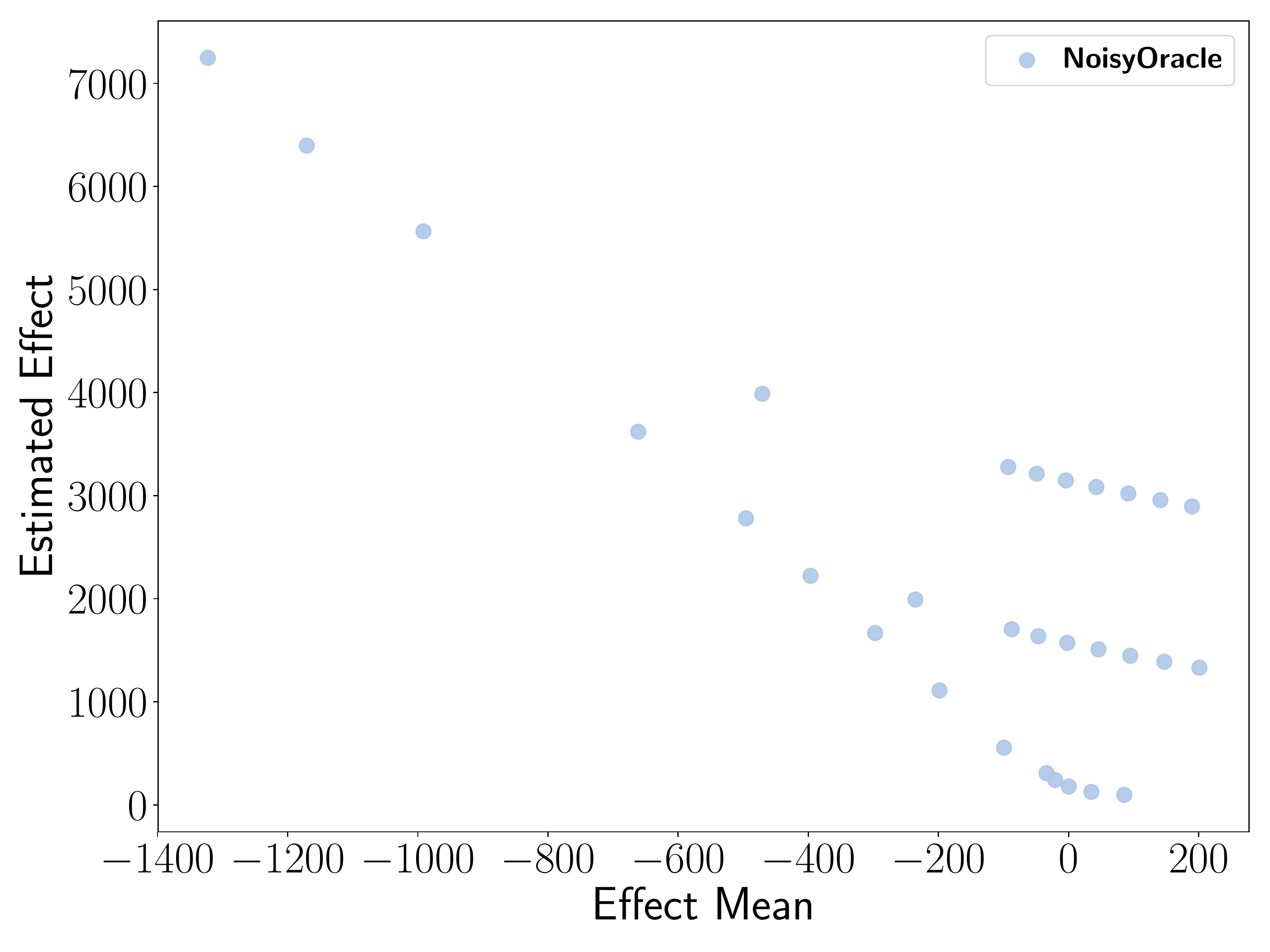}}
    \subfloat[]{\includegraphics[width=0.3\textwidth]{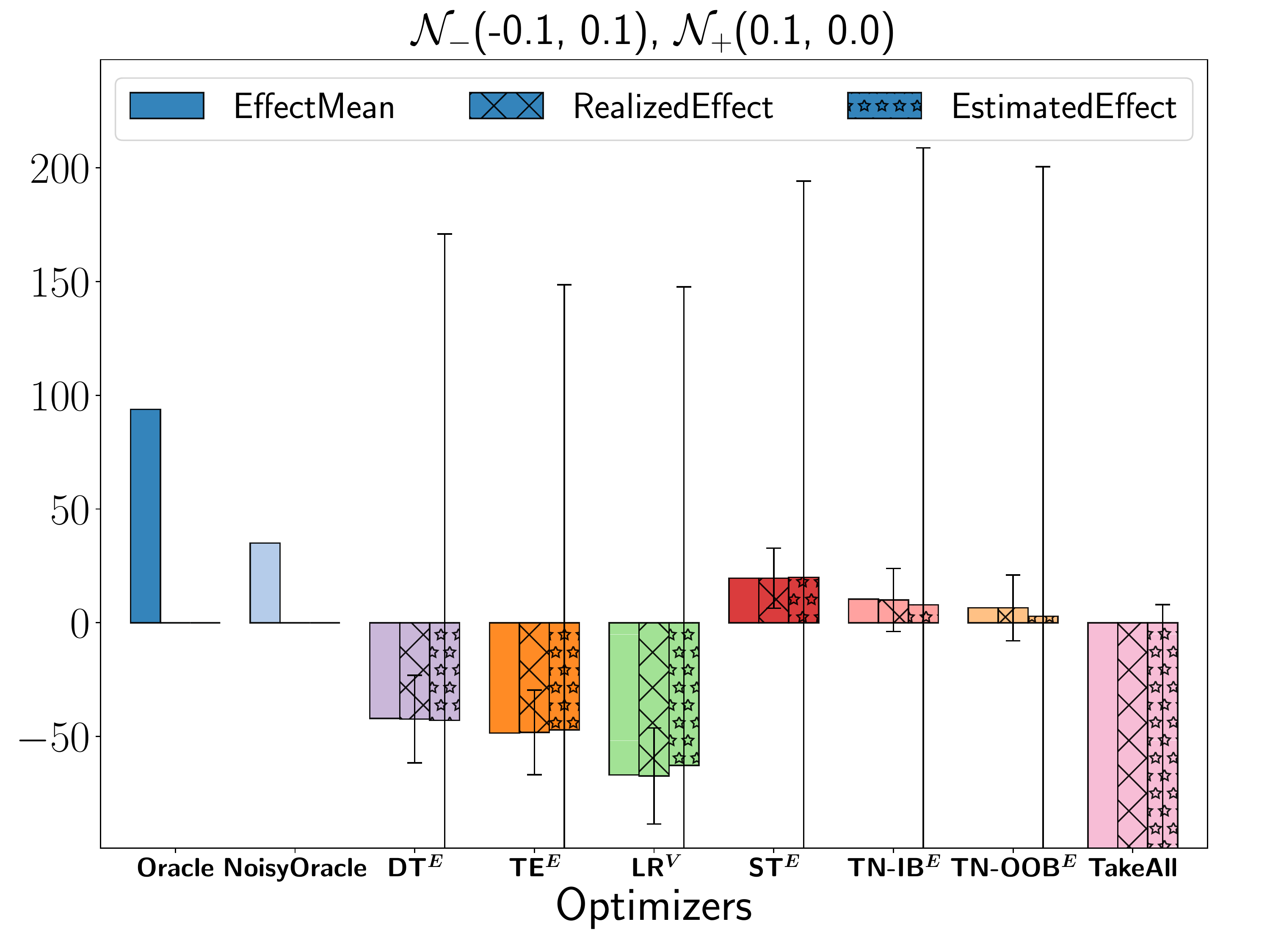}}

    \subfloat[]{\includegraphics[width=0.3\textwidth]{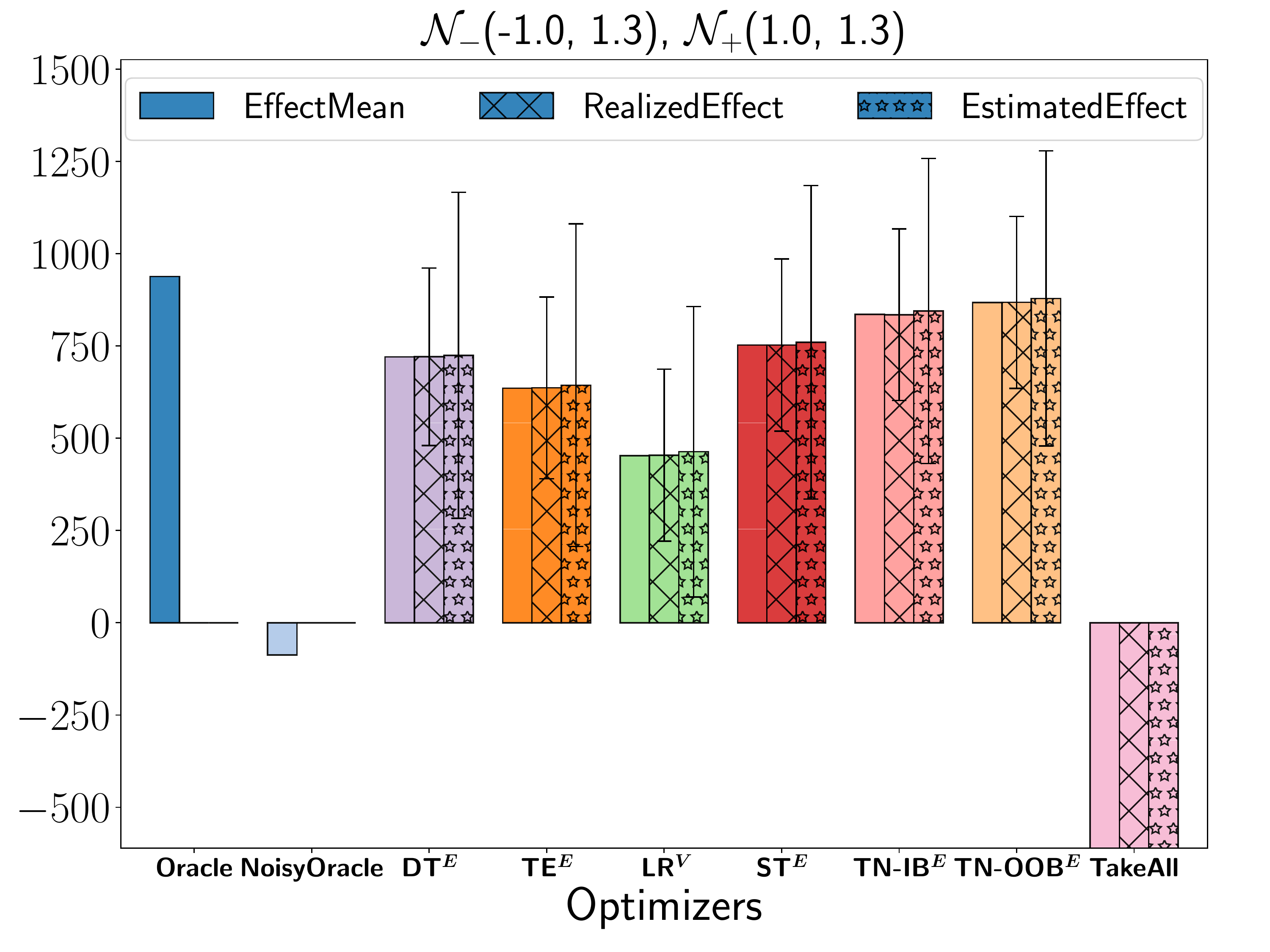}}
    \subfloat[]{\includegraphics[width=0.3\textwidth]{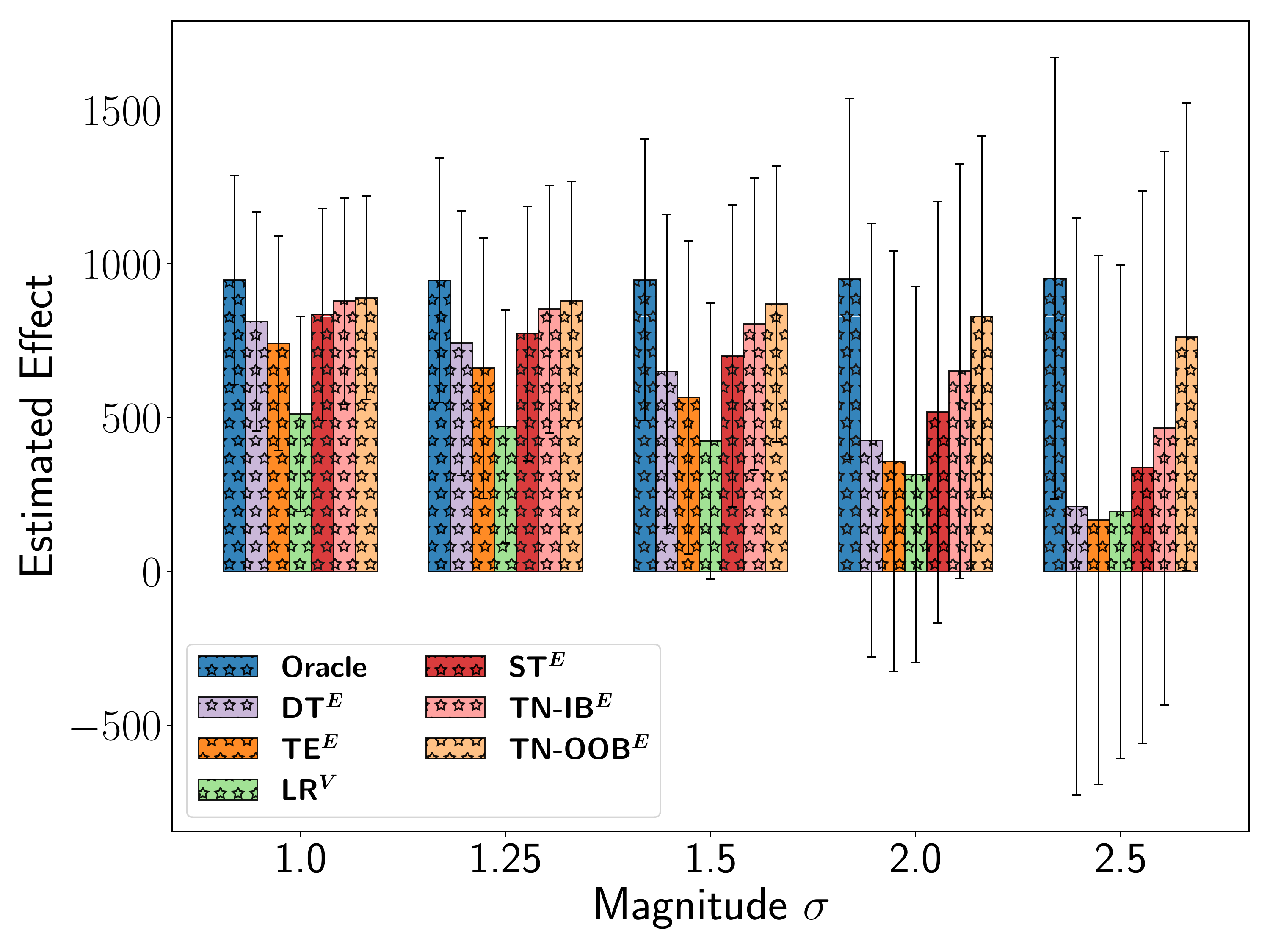}}
    \subfloat[]{\includegraphics[width=0.3\textwidth]{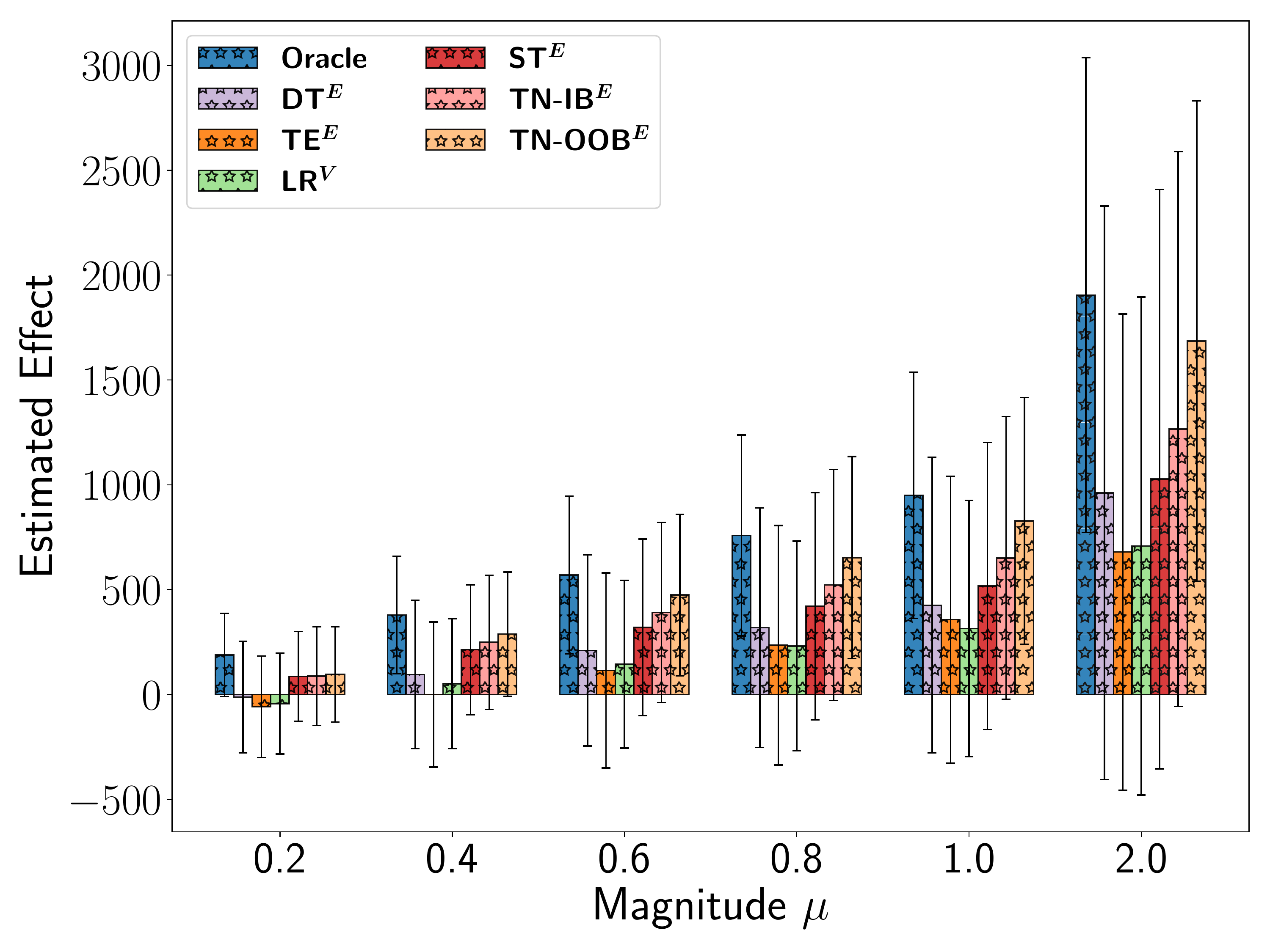}}

    \caption{(a) Unbiased effect estimation.  (b) Bias of NoisyOracle selection criteria.   Estimation of (c) small and (d) moderate treatment effects. Effect of varying (e) $\mu$ and (f) $\sigma$ on estimation.}
    \label{fig:synthetic_boston}
\end{figure}

\subsubsection{Policy Evaluations and Optimization}
We first show that the proposed estimators are unbiased: \figref{fig:synthetic_boston}a shows the complete range of learners, as prescribed by \secref{sec:optimizers}, crossed with the complete range of tested $\mu_+, \mu_{-}, \sigma_+, \sigma_{-}$ treatment effects (with $2 \geq |{\mu_+}|,|{\mu_{-}}| \geq 0.1$ and $5 \geq |{\sigma_+}|,|{\sigma_{-}}| \geq 0$), as prescribed by \secref{sec:simulatedeffectdistributions}.  The x-axis shows the true average effect of the chosen policy, while the y-axis shows the Estimated Effect by each learner, and we see that these are nearly perfectly correlated with their true Effect Mean.  In contrast, when applied on \textbf{NoisyOracle}, the effect estimator frequently exhibits large bias in comparison to the true average mean \figref{fig:synthetic_boston}b.  This is because policies produced by \textbf{NoisyOracle} violate the conditions of Theorem \ref{thrm:unbiasedness}.

Next, we examine specific small ($\mu,\sigma \approx .1$) and moderate ($\mu,\sigma \approx 1$) treatment effects in \figref{fig:synthetic_boston}c-d.  For brevity we examine a subset of learners; additionally, we clip the result of \textbf{TakeAll} (which is strongly negative) to focus on the other methods.  We note that the Estimated Effect column closely matches the EffectMean column, indicating that the effect estimator is indeed unbiased.  For the small treatment effect we note that although the estimator noise dominates the signal, on average, all the optimizers are detecting some signal, and some are producing positive value.  For the moderate treatment effect (\figref{fig:synthetic_boston}d), all of the proposed methods show significant gain over both the biased \textbf{NoisyOracle} as well as choosing no samples.

In \figref{fig:synthetic_boston}e, we fix $|\mu_+|,|\mu_{-}| = 1$ and vary the amount of noise $\sigma$. We drop the biased \textbf{NoisyOracle} and poorly performing \textbf{TakeAll} to focus on the Estimated Effect of each learner. We additionally report the Estimated Effect of \textbf{Oracle} as a baseline of maximum possible performance.  For $\sigma=1$ we find all of the methods perform well, many with an expected policy near that of the \textbf{Oracle}.  For many methods as $\sigma>2$ we see some additional variance; however, \textbf{TN-OOB}$^E$ remains statistically significant.  Lastly, in \figref{fig:synthetic_boston}f, we fix the ratio $\nicefrac{|\sigma|}{|\mu|}=2$ and vary the magnitude of the negative and positive $\mu$ values.  We find the \textbf{TN-OOB}$^E$ is the best performer, producing policies that are statistically provable to provide positive values even at this low signal to noise level, for $\mu$ values 0.4 and higher.

\section{Real world experiments}
\label{sec:real}

We provide some statistics of applying the methods described here to three real world experiments ran by a large internet service on business customers.  In all three experiments, the target variable is revenue.  Unlike A/B testing on end-users, experiments on business customers are uncommon in the industry, as they are much harder to run and analyse.  Dataset sizes are much smaller (in the 1Ks to 10Ks, v.s. up to 100Ms for end users), skew in interesting measures, such as revenue is very large.  It also appears that response times are very long, and hence experiments must be run for months before signal can be detected.

Table \ref{tab:real_exp} provides statistics for the three experiments.  The first observation to take is the extremely high levels of noise, as seen in the \textbf{STE} of the \textbf{IP} estimator.  This is the standard error of the effect estimator, and is the magnitude of effect one needs to achieve to get a z-score of 1, i.e. a p-value of 0.16.  The \textbf{DR} estimator achieves better noise levels, in all three datasets, and is the estimator we refer to in the rest of this discussion.  Next note that whereas for the general population, Experiment 1 causes a 27\% reduction in revenue, the optimizer found a policy with no reduction in revenue.  Indeed, it appears that almost all the reduction was due to approximately 5\% of the customers (not shown), and the optimizer excluded those from the policy.  However no gain in revenue could be shown for the policy.  Experiments 2 and 3 show no statistically significant result, illustrating the hardness of the task.  Nevertheless, in both cases there is some evidence that the optimized policy outperforms taking all or no samples (both \textbf{Optimized} and \textbf{Opt-All} estimates are positive).

A note about the p-value calculation: We calculate Fisher's Exact P-Values for rejecting a sharp-null-hypothesis of no-effect \cite{fisher1935design,imbens_rubin_2015_c5}. To this end, we randomly draw 1,000 fake complete treatment indicator vectors, and for each of those run the optimizer followed by the effect estimator, generating a single (fake) estimate from each draw.  The p-value is the fraction of fake estimates that are higher than the estimate for the real data.

\begin{table}
\begin{tabular}{llllllllll}
\toprule
             & Policy & \multicolumn{2}{l}{-} & \multicolumn{2}{l}{TakeAll} & \multicolumn{2}{l}{Optimized} & \multicolumn{2}{l}{Opt-All} \\
             & Estimate &      N &    M &      IP &     DR &        IP &     DR &      IP &     DR \\
\midrule
Experiment 1 & Value &  21114 &  197 &    -36\% &   -27\% &      -10\% &    -0\% &     27\% &    27\% \\
             & STE &        &      &     25\% &    16\% &           &        &         &        \\
             & PVal &        &      &   0.934 &  0.958 &     0.703 &  0.490 &   0.105 &  0.047 \\
Experiment 2 & Value &   1461 &   73 &      9\% &     6\% &       13\% &     8\% &      3\% &     2\% \\
             & STE &        &      &     17\% &    10\% &           &        &         &        \\
             & PVal &        &      &   0.303 &  0.276 &     0.182 &  0.138 &   0.409 &  0.411 \\
Experiment 3 & Value &   6077 &  792 &      6\% &     0\% &        1\% &     2\% &     -6\% &     2\% \\
             & STE &        &      &     12\% &     5\% &           &        &         &        \\
             & PVal &        &      &   0.222 &  0.494 &     0.499 &  0.235 &   0.771 &  0.240 \\
\bottomrule
\end{tabular}

\caption{Statistics from Three Real World Experiments}
\caption*{Effect estimates and other statistics are provided for applying the treatment to all samples (\textbf{TakeAll}), for the policy found by our optimizer (\textbf{Optimized}), and for the difference between \textbf{Optimized} and \textbf{TakeAll} (\textbf{Opt-All}).  Estimate are produced by either inverse-propensity-weighting (\textbf{IP}) or doubly-robust estimator (\textbf{DR}).  \textbf{N}: number of samples; \textbf{M}: number of features; \textbf{Value} The effect estimate, given in percentage of total measurement;  \textbf{STE} Standard error of the effect;  \textbf{PVal} p-value for estimate  }
\label{tab:real_exp}
\end{table}

\section{Conclusion}
\label{sec:conclusion}
In this paper, we presented a bagging procedure for learning optimized policies and getting an unbiased estimate of the value of deploying them.  We presented sufficient conditions for providing an unbiased estimation for the value of an optimized policy and proved the bagging procedure meets the criteria.  We conducted thorough simulation studies on standard public datasets and empirically demonstrated that our procedure is an unbiased estimator.  Lastly, we demonstrated on our same methods to real world trials and demonstrated its power to detect positive policies even when the average treatment effect is negative.

Future work includes characterising and bounding the variance, with ideas described in \cite{wager2017estimation} likely to be useful in bounding the correlation between $T_i$ and $\Pi_j$ and thus helping to bound the variance of the estimator.  On the practical side, we believe we can see gains by improving our $L^{\left(0\right)}$ and $L^{\left(1\right)}$ trainers, and indeed in using separate ones for the treatment and the control.  We also consider applying the methodology to the more classical end-user A/B testing experiments, where the noise level is significantly lower.

\section{Acknowledgments}
\label{sec:acknowledgments}

The authors thank their colleagues in Microsoft Bing Ads for their help.  In particular, Ajith Moparthi and Liangzhong Yin for support in running experiments, Wenhao Lu and Amjad Abu-Jbara for providing their experiments' data,  Michal Prussak for his help editing the manuscript, and all of the above as well as Neil P. Slagle and Patrick R. Jordan for helpful discussions.

\clearpage
\bibliography{ms}
\bibliographystyle{abbrv}



\end{document}